\documentclass[sigconf,nonacm]{acmart}
\usepackage{algorithm}
\usepackage{algorithmic}

\usepackage{textcomp}
\usepackage{multirow}
\usepackage{colortbl}
\definecolor{mygray}{gray}{.88}
\usepackage{xcolor}
\usepackage{verbatim}
\usepackage{appendix}
\usepackage{bbding}
\usepackage{listings}
\usepackage{subcaption}
\usepackage{pifont} 

\renewcommand\footnotetextcopyrightpermission[1]{}
\begin{document}

\title{CMLCompiler: A Unified Compiler for Classical Machine Learning}
\author{Xu Wen}
\affiliation{%
  \institution{Institute of Computing Technology, Chinese Academy of Sciences \and University of Chinese Academy of Sciences}
  \city{Beijing}
  \country{China}
}
\email{wenxu@ict.ac.cn}

\author{Wanling Gao}
\affiliation{%
  \institution{Institute of Computing Technology, Chinese Academy of Sciences \and University of Chinese Academy of Sciences}
  \city{Beijing}
  \country{China}
}
\email{gaowanling@ict.ac.cn}

\author{Anzheng Li}
\affiliation{%
  \institution{Institute of Computing Technology, Chinese Academy of Sciences \and University of Chinese Academy of Sciences}
  \city{Beijing}
  \country{China}
}
\email{lianzheng20g@ict.ac.cn}

\author{Lei Wang}
\affiliation{%
  \institution{Institute of Computing Technology, Chinese Academy of Sciences \and University of Chinese Academy of Sciences}
  \city{Beijing}
  \country{China}
}
\email{wanglei_2011@ict.ac.cn}

\author{Zihan Jiang}
\affiliation{%
  \institution{Huawei Technologies Co., Ltd.}
  \city{Beijing}
  \country{China}
}
\email{jiangzihan0512@gmail.com}

\author{Jianfeng Zhan}
\authornote{Corresponding author.}
\affiliation{%
  \institution{Institute of Computing Technology, Chinese Academy of Sciences \and University of Chinese Academy of Sciences}
  \city{Beijing}
  \country{China}
}
\email{zhanjianfeng@ict.ac.cn}
\renewcommand{\shortauthors}{Xu Wen, Wanling Gao, Anzheng Li, Lei Wang, Zihan Jiang, and Jianfeng Zhan}

\begin{abstract}
Classical machine learning (CML) occupies nearly half of machine learning pipelines in production applications. Unfortunately, it fails to utilize the state-of-the-practice devices fully and performs poorly.
Without a unified framework, the hybrid deployments of deep learning (DL) and CML also suffer from severe performance and portability issues.
This paper presents the design of a unified compiler, called CMLCompiler, for CML inference. We propose two unified abstractions: operator representations and extended computational graphs. The CMLCompiler framework performs the conversion and graph optimization based on two unified abstractions, then outputs an optimized computational graph to DL compilers or frameworks. We implement CMLCompiler on TVM. The evaluation shows CMLCompiler's portability and superior performance. It achieves up to 4.38$\times$ speedup on CPU, 3.31$\times$ speedup on GPU, and 5.09$\times$ speedup on IoT devices, compared to the state-of-the-art solutions --- scikit-learn, intel sklearn, and hummingbird. 
Our performance of CML and DL mixed pipelines achieves up to 3.04x speedup compared with cross-framework implementations. 
The project documents and source code are available at \url{https://www.computercouncil.org/cmlcompiler}.
\end{abstract}



\maketitle

\section{Introduction}

Deep learning (DL) and classical machine learning (CML), collectively called machine learning (ML), have played an increasingly critical role in recent years. 
DL refers to those neural network models, such as convolutional neural networks (CNNs)~\cite{li2021survey}, recurrent neural networks (RNNs)~\cite{medsker1999recurrent}, and generative adversarial networks (GANs)~\cite{goodfellow2014generative}.
Different from DL, CML represents a set of non-neural network models in ML, e.g., linear models~\cite{searle2016linear}, decision trees~\cite{loh2011classification}, random forests~\cite{breiman2001random}, and support vector machines~\cite{suthaharan2016support}.
DL stands out because of its accuracy, while CML is still widely used for lower time and energy costs. 
Doris Xin et al.~\cite{xin2021production} analyze 3000 production ML pipelines at Google and find that 40\% of them use CML models. 
Besides, many real-world applications adopt hybrid deployments of CML and DL~\cite{amazonsegamker} to guarantee high accuracy and low latency~\cite{reimers2019sentence,ma2019universal,sengupta2020nuclear,ling2017model}, e.g., DL models for feature embedding and CML models for classification or regression.
\begin{figure}
	\centering
	\includegraphics[width=0.48\textwidth]{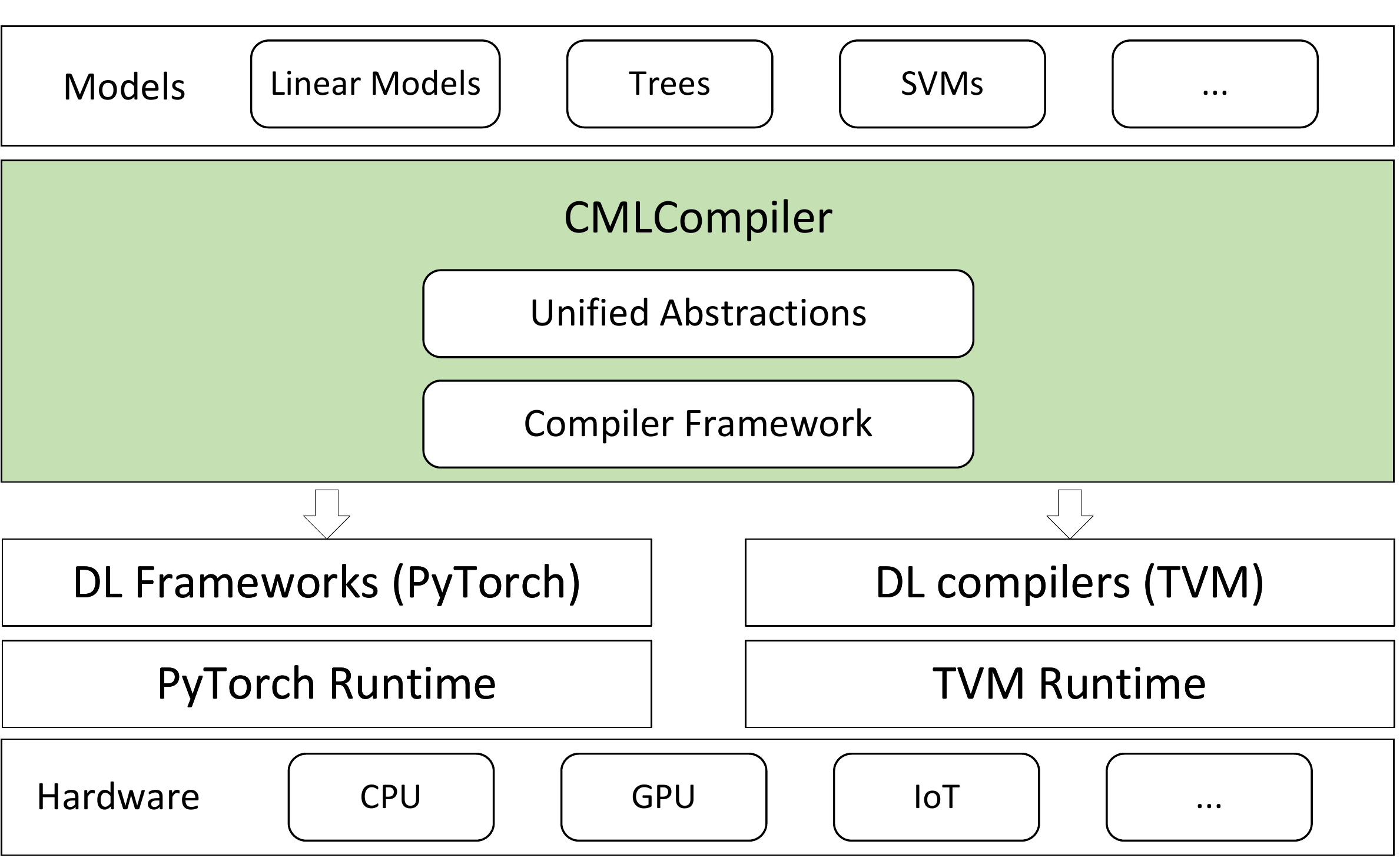}\\
	\caption{
		The CMLCompiler design. Our contributions are highlighted in green color.
	}
	\label{framework_introduction}
\end{figure}

DL compilers, like TVM~\cite{10.5555/3291168.3291211,DBLP:journals/corr/abs-1801-08058, lattner2020mlir}, provide a structural approach to tackle the portability issue and facilitates wide deployment of DL models on a broad spectrum of devices like GPUs, FPGAs, and IoT devices and guarantees an appreciable performance. 
DL compilers use computational graphs as high-level abstractions, supporting a large variety of DL models. Meanwhile, DL compilers propose low-level abstractions such as tensor representation to generate executable code. For newborn hardware, the vendor just needs to provide hardware primitives, instead of a sophisticated high-performance library that is prohibitively costly. Based on the tensor representation and computational graphs abstractions, many optimizations~\cite{chen2018learning, jia2019taso, zheng2020ansor}  are proposed to boost performance, e.g., they provide sophisticated support for CPU processor architectures as the latter has different architectures, diverse core numbers, extended instructions, and cache sizes.

However, despite its popularity and importance, CML suffers from severe portability and performance issues.
State-of-the-practice and state-of-the-art CML frameworks~\cite{10.5555/1953048.2078195, 10.5555/2946645.2946679, h2o_platform}  provide ad-hoc solutions,  implementing each CML model on every hardware device case by case due to the lack of unified abstractions. These ad-hoc solutions raise considerable difficulties in developing a general-purpose framework and optimization techniques to achieve optimal performance for every model. 
They either lack the support or only partially support various hardware devices, such as GPUs, FPGAs, and IoT devices. In addition, adding support for a model on a new hardware device needs great effort, more than several thousands of lines of codes~\cite{cudatree}, let alone hundreds or thousands of models and devices. Moreover, they also face performance issues. Even on the CPUs -- the most popular CML platform, the performance is unsatisfactory due to the lack of specific optimizations for advanced characteristics like multi-cores and SIMD.
The hybrid deployment of CML and DL models faces more severe problems.

Our intuition is to enable CML to leverage DL's well-defined unified abstractions and highly mature compilers,  optimization technologies, and frameworks. Unfortunately, it is not a trivial task. There are significant distinctions in operators and models between CML and  DL. 
DL operators focus on tensors, while CML handles arrays, matrices, scalars, and tables.
DL models are all neural network models, while CML models, such as decision trees, can hardly be represented as neural networks.
Most DL models are expressible as flat sequences of operations without if-statements~\cite{reed2022torch}, but if-statements frequently occur in CML models.
Existing DL abstractions, such as tensor representation and computational graphs, can not directly represent CML operators and models.
Those distinctions determine CML can hardly leverage the DL ecosystems directly. Several efforts attempt to support CML models on DL frameworks, e.g., TensorFlow~\cite{10.5555/3026877.3026899} provides a CPU-based decision forest library TF-DF~\cite{tf_df}. 
However, these attempts do not solve the generality and portability issue. They only support a narrower range of models, lacking support for GPUs and IoT devices.

This paper focuses on CML inference for the first step, considering its great significance that occupies nearly half of the total cost~\cite{amazonsegamker} and its wide applications in online serving, Internet of things (IoT), etc~\cite{8327042,8675201}. We will extend our work to CML training in the near future. 
As illustrated in Fig.~\ref{framework_introduction}, we propose a unified compiler,  CMLCompiler, for CML inference, which enables CML to leverage the mature DL ecosystems. At the core of CMLCompiler are two unified abstractions: operator representations and extended computational graphs (ECGs) and a compiler framework. 
Operator representations convert CML operators into tensor formats, while an ECG organizes these converted operators in an optimization-friendly way. The two unified abstractions define how to convert and translate CML models into DL computational graphs, which can be recognized and executed by DL frameworks and compilers. The CMLCompiler framework consists of four modules -- operator converter, model parser, graph optimizer, and graph translator.
The CMLCompiler framework performs the conversion and graph optimization based on two unified abstractions, then outputs an optimized DL computational graph to DL compilers or frameworks. 
CMLCompiler can also optimize the mixed pipelines of CML and DL.
As TVM provides portability and sophisticated optimizations, we choose to implement CMLCompiler on TVM. Currently, it supports up to 35 CML models.

This paper makes the following contributions:
\begin{itemize}
    \item We propose two unified abstractions -- operator representations and extended computational graphs-- to represent CML operators and models.
    \item We present the design of CMLCompiler, a unified compiler for CML inference, based on these abstractions. 
    The CMLCompiler framework performs the conversion and graph optimization based on two unified abstractions, then outputs an optimized DL computational graph to DL compilers or frameworks. 
    \item CMLCompiler enables the hybrid deployment of CML and DL with a unified framework.
    \item We implement CMLCompiler on top of TVM, achieving up to 4.38x speedup on CPU, 3.31x speedup on GPU, and 5.09x speedup on IoT devices, compared to the state-of-the-art solutions --- scikit-learn, intel sklearn, and hummingbird. Our support for CML and DL mixed pipelines achieves up to 3.04x speedup compared with cross-framework implementations.
\end{itemize}

The remainder of the paper is organized as follows.
Section~\ref{section_background} introduces the motivation.
Section~\ref{section_abstraction} introduces unified abstractions.
Section~\ref{section_design} shows design and implementation.
Section~\ref{section_evaluation} presents our evaluation.
Section~\ref{section_related_work} illustrates the related work. Finally, we draw a conclusion in Section~\ref{section_conclusion}.

\section{Motivation}\label{section_background}
\begin{figure}
	\centering
	\includegraphics[width=0.48\textwidth]{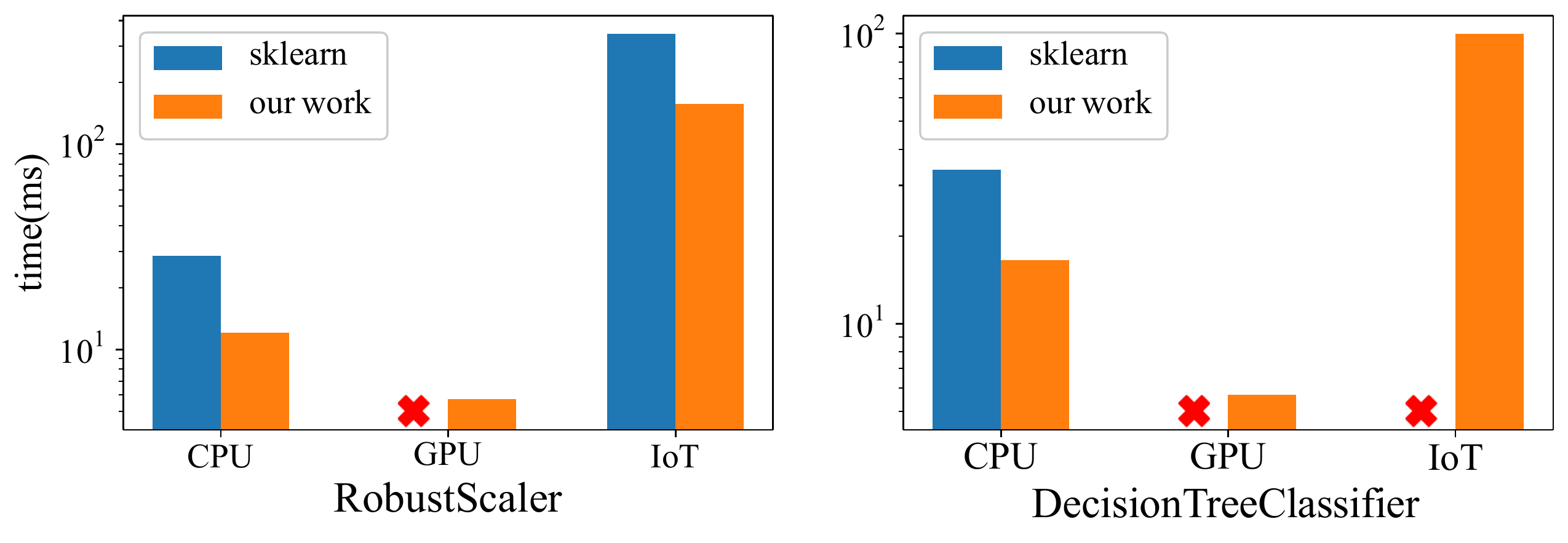}\\
	\caption{This figure compares the performance of sklearn, the most widely used CML framework on GitHub~\cite{DBLP:journals/corr/abs-1912-09536}--- against CMLCompiler.
		Our evaluation shows that sklearn suffers from both performance and portability issues  for a lack of unified abstractions.
            "$\times$" means unsupported.
	}
	\label{motivation}
\end{figure}

CML faces severe portability and performance issues.
Fig.~\ref{motivation} compares the performance of sklearn, the most widely used CML framework on GitHub~\cite{DBLP:journals/corr/abs-1912-09536}--- against CMLCompiler leveraging DL compilers. 
We find that sklearn can not support GPUs and only supports IoT devices partially.
Adding support for a new hardware device needs great effort due to the ad-hoc implementations.
For example, adding support for random forest on GPU needs 2.7k lines of code~\cite{cudatree}.
Many models and hardware devices need to be supported, requiring hundreds or thousands of more effort. 
Moreover, due to the lack of compilation support for CPU features,  sklearn has poor performance. 
As shown in Fig.~\ref{motivation}, CMLCompiler achieves 2.3x speedup by utilizing AVX2 through compilation compared with sklearn.  
Other CML frameworks such as Spark MLlib~\cite{10.5555/2946645.2946679} and H2O~\cite{h2o_platform} face the same problems.
Our solution is to propose unified abstractions to utilize DL compilers and frameworks, achieving portability and high performance. 

CML and DL models are often deployed hybrid in NLP~\cite{reimers2019sentence}, intelligent healthcare~\cite{sengupta2020nuclear}, recommendation systems~\cite{ling2017model}, etc., especially in scenarios with limited computational power and small datasets.
Many of them are deployed on heterogeneous hardware devices for online serving.
As there is no unified system, different frameworks are deployed with three disadvantages.
First, this limits the portability.
If one framework fails on the target device, the whole pipeline corrupts.
Second, there are extra costs due to data conversions across frameworks.
Third, it is hard to make optimizations across different frameworks.
Using a unified framework can overcome these disadvantages, so we add the support for hybrid deployment of CML and DL in CMLCompiler.

\section{The Unified Abstractions}\label{section_abstraction}
CMLCompiler takes CML models as input and returns DL computational graphs as output, utilizing DL frameworks or compilers to compile and deploy them.
At the core of CMLCompiler are two unified abstractions.
Operator representations are used to represent CML operators in tensor format, as shown in Section~\ref{operator_representation}.
Extend computational graph (ECG) organizes operator representations in an optimization-friendly way and can be used to represent CML models, as shown in Section~\ref{section_ECG}.
Section~\ref{section_supported} shows the supported algorithms and extensions for other algorithms.
\subsection{Operator Representation}\label{operator_representation}

An operator representation uses a combination of one or more DL operators with tensors as input and output to represent a CML operator.
We convert CML operators into DL operators and wrap them in the format of operator representations.
Data in CML has mainly four formats: arrays, matrices, scalars, and tables~\cite{vanderplas2016python}.
Matrices and arrays are regarded as two types of tensors whose operators can naturally be converted into DL operators.
When CML models deal with tables, they take numeric data from tables and operate it, which can also be regarded as scalars.
Hereby, we focus on the operators on scalars.
\subsubsection{Operator categories and corresponding representations.}\label{section_conversion_category}

\begin{table*}
	\caption{
	The summary of operator representation. Each operator representation represents a CML operator.
	Scalars are marked as lower-case letters, while tensors are marked as upper-case letters.
	EW is short for element-wise.
	}
	\label{operator_type}
	\center
	{
		\begin{tabular}
			{c|c|c|c}
			\hline
			\multicolumn{2}{c|}{CML operators in scalar format} & \multicolumn{2}{c}{Operator Representation in tensor format}\\
			\hline
			Operator Type & Expressions & Operator Type &Expressions\\
			\hline
			Assignment &$x_1 \gets v_1;\ x_2 \gets v_2;\ ...;\ x_n \gets v_n$ & Assignment & $X = [x_1,x_2,...,x_n];\ V = [v_1,v_2,...,v_n];\ X \gets V$ \\
			\hline
			Swap & $x_1 \gets x_2;\ x_2 \gets x_1;$ &Reorganization & $X = [x_1,x_2];\ reshape(X);$\\
			\hline
			Basic Arithmetic &$x_1 + y_1;\ x_2 + y_2;\ ...;\ x_n + y_n$  &EW Arithmetic & $X = [x_1,x_2,...,x_n];\ Y = [y_1,y_2,...,y_n];\ X + Y$  \\
			\hline
			Aggregation & $sum(x_1,x_2,...,x_n)$ & Reduction & $X = [x_1,x_2,...,x_n];\ sum(X)$\\
			\hline
			Comparison& $x_1 < y_1;\ x_2 < y_2;\ ...;\ x_n < y_n$ &EW Comparison & $X = [x_1,x_2,...,x_n];\ Y = [y_1,y_2,...,y_n];\ X < Y$\\
			\hline
			Conditional& $if(expr1)\ expr2\ else\ expr3$& \multicolumn{2}{c}{
			Described in Section~\ref{section_conditional_operator}}\\
			\hline
	\end{tabular}}
\end{table*}

As shown in Table~\ref{operator_type}, we classify CML operators into six categories and provide operator representations, respectively.

(1) Assignment operators assign values to variables. 
If we assign n values $v_1,\ v_2,\ ...,\ v_n$ to n variables $x_1,\ x_2,\ ...,\ x_n$, we organize these variables and values in two tensors $X = [x_1,x_2,...,x_n]$ and $V = [v_1,v_2,...,v_n]$.
Then we assign tensor V to tensor X to replace n scalar assignments.
Tensor assignments benefit memory copy which stores data in block.

(2) Swap operators swap two or more variables.
These variables can be represented in a tensor format and use reorganization operators such as $reshape$ to swap the elements.

(3) Basic arithmetic operators refer to those arithmetic calculations based on scalars, such as $add$, $sub$, $mul$, and $div$.
We use element-wise arithmetic operators based on tensors to replace them, which can utilize SIMD instructions better.

(4) Aggregation operators refer to operators that calculate aggregates among many scalars, such as $min$, $max$, $sum$, and $avg$.
Reduction operators can be used to accomplish that.

(5) Comparison operators make a comparison between scalars and return True or False, such as $less$, $equal$, and $greater$.
Comparisons with the same operator can be represented in a tensor format and use an element-wise comparison to replace.

(6) Conditional operators are used to represent if-else statements, in the form of $if(expr1)\ expr2\ else\ expr3$, where $expr1$ is a comparison operator.
If $expr2$ and $expr3$ are all assignment or arithmetic operators, we convert all three expressions into tensors.
However, the situation gets tricky if one of $expr2$ or $expr3$ is still a conditional operator.
We call those operators sequential conditional operators.
Sequential conditional operators may contain many conditions, where each element in a tensor may have quite different decision paths. 
The complexity of decision paths makes it difficult to convert those operators into tensor operators. 
Those frequent if-else statements perform poorly on hardware devices such as GPUs and ASICs.
Sequential conditional operators are the most delicate, and we defer their discussion later.

\subsubsection{Conditional operators representation.}\label{section_conditional_operator}

\begin{figure*}
	\centering
	\includegraphics[width=1.0\textwidth]{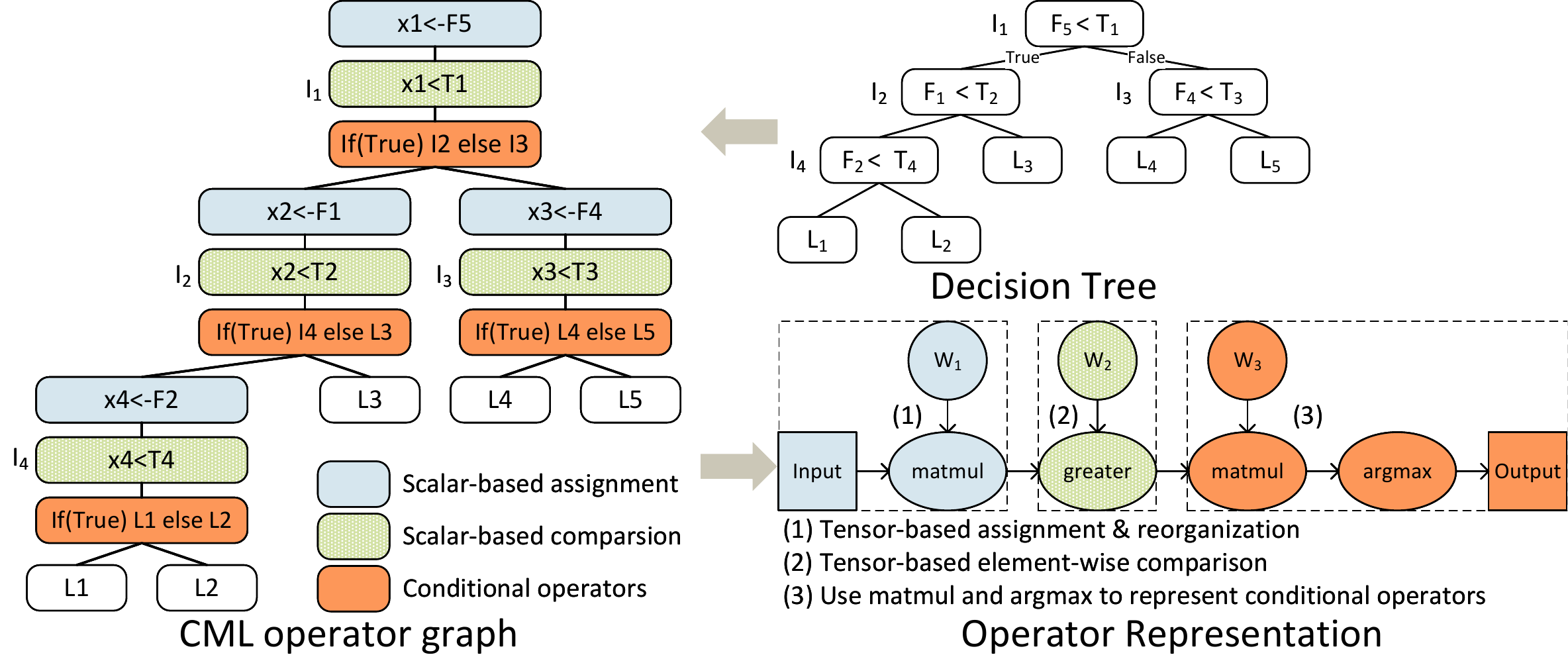}
	\caption{Converting Decision Tree to operator representations. 
            Sequential conditional operators are represented by the combination of $matmul$ and $argmax$, $matmul$ is short for matrix multiplication.
		$F$, $T$, $I$, and $L$ refer to features, thresholds, internal nodes, and leaf nodes.
		$W_1$, $W_2$, and $W_3$ are the weights of DL operators, whose definitions and properties are shown in Table~\ref{weight_property}.
	}
	\label{conversion}
\end{figure*}
\begin{table}
	\caption{The properties of weights in Fig.~\ref{conversion}. 
	$N_S$, $N_F$, $N_I$, and $N_L$ refer to the number of samples, features, internal nodes, and leaf nodes, respectively.
	$Input\in\mathbb{R}^{N_S \times N_F}$ means $N_S$ samples, each has $N_F$ features.
	$W_1\in\{0,1\}^{N_F \times N_I}$ captures the relationship between features and internal nodes.
	$W_2\in\mathbb{R}^{N_I}$ is the thresholds used in internal nodes.
	$W_3\in\{0,1\}^{N_I \times N_L}$ represents the structure between internal nodes and leaf nodes.
	$Output \in\mathbb{N}^{N_S}$ returns the leaf node index each sample reaches.
	Dtype is the data type of weights. 
	Sparsity is the ratio of non-zero data to all data in weights.
	}
	\label{weight_property}
	\center
	{
        \begin{tabular}{*{3}{c}}
          \toprule
          Definition &Dtype &Sparsity \\
          \midrule
           $W_1[i][j]=
           \left\{\begin{tabular}{@{\ }l@{}}
            $1, F_i \in Condition(I_j)$ \\
            0, otherwise  
          \end{tabular}\right.$ 
          & bool& $\frac{1}{N_F}$\\
          \midrule
          $W_2[i]=Threshold(I_i)$
          &float32&1\\
          \midrule
           $W_3[i][j]=
           \left\{\begin{tabular}{@{\ }l@{}}
            $0, L_j\in LeftSubTree(I_i)$
            \\ 1, otherwise
          \end{tabular}\right.$ 
          & bool & $[\frac{1}{2}, 1-\frac{1}{N_L}]$\\          
          \bottomrule
        \end{tabular}
    }
\end{table}
We analyze those widely used CML models and find that sequential conditional operators mainly occur in tree-based models.
So we use decision tree as an example to introduce the representation of conditional operators in detail, as shown in Fig.~\ref{conversion}.
We use the combination of DL operators to represent those sequential conditional operators.

The top-right shows the original decision tree.
The input data is a list of samples; each has many features.
$I$ refers to internal nodes, numbered in the order of Level Order Traversal.
$L$ refers to leaf nodes, numbered in the order of In-Order Traversal.
Each leaf node is an assignment operator, reaching which node determines the final result.
The left shows the corresponding CML operator graph.
Scalar-based assignments assign features $F_j$ to temporary variables $x_i$, scalar-based comparisons compare $x_i$ with thresholds $T_i$, and conditional operators determine the next nodes to be reached.
These scalar-based CML operators are converted to operator representations, as shown in the bottom-right in Fig.~\ref{conversion}, and the definitions and properties of weights are shown in Table~\ref{weight_property}.
Multi Scalar-based assignments are converted to one Tensor-based assignment. 
((1) in Fig.~\ref{conversion})
The data processing order differs from its layout in the tensor, we use the $reorganization$ operator to change the tensor layout, which is replaced by $matmul$ with 0-1 matrix $W_1$. 
((1) in Fig.~\ref{conversion})
Multi Scalar-based comparisons are converted to one Tensor-based element-wise comparison. 
((2) in Fig.~\ref{conversion})
Sequential conditional operators are represented by the combination of $matmul$ and $argmax$. 
((3) in Fig.~\ref{conversion})
Output returns the leaf nodes every sample reaches.
Finally, a decision tree is converted to the combination of three operator representations.

\subsubsection{The features of CML operator representations.}\label{section_feature}
As described above, we represent CML operators in the format of operator representations.
These operator representations have unique features different from operators in DL models.

First, the weights of DL operators and CML operator representations have different meanings.
The weights in DL models are all learnable parameters.
Without approximate optimizations such as pruning and quantization, those weights are dense, and the data type (dtype) should be float32 to ensure accuracy.
Many weights of CML operator representations have other meanings, such as representing the structure of conditional operators.
Those weights are sparse and can naturally be expressed as low-precision dtypes such as bool.
The natural sparse features bring optimizations described in Section~\ref{section_sparse_operator_replacing}.

Second, the frequent operators in DL and CML are not the same.
Almost all operators in DL take float32 as input and return float32 as output.
CML uses many comparison operators, such as $less$, $equal$, and $greater$, which rarely occur in DL models.
Those comparison operators take float or integer as input and return bool, bringing remarkable changes in the dtype of input and output, which can be used to make optimizations as described in Section~\ref{section_dtype_rewriting}.
Both DL and CML models use indices operators, which compare input and returns indices, such as $argsort$ and $argmax$.
Those indices operators have mathematical properties that can be used to make graph-level optimizations, as described in Section~\ref{section_redundant_elimination}.
These optimizations can be ignored in DL models with dozens or hundreds of layers but are helpful for those CML models with fewer layers.
\subsection{Extended Computational Graph}\label{section_ECG}
This section introduces extended computational graph (ECG), which organizes operator representations in an optimization-friendly way and can be used to represent CML models.
ECG is an extension based on DL computational graph.
In general, a DL computational graph is represented as a directed graph where nodes represent operations on tensors or program inputs and edges represent data dependencies between operations~\cite{10.5555/3291168.3291211}.
From the perspective of the DL frameworks and compilers, computational graphs are dense and float32 by default, such as neural network models. Using approximate optimizations like pruning and quantization brings sparse and low-precision data to all operators and weights. These optimizations cause a decrease in accuracy and bring extra computation, such as calibration.
When we convert CML operators to operator representations, part of those converted operators and weights are sparse and low-precision naturally. 
Using DL computational graphs to represent CML models directly is not precise enough and ignores many optimization opportunities due to the data type and sparse features.
So we extend the computational graph in the DL systems into extended computational graph (ECG) as the unified abstraction for CML models.

Before introducing ECG, first, we present more details about data type (dtype) and sparsity.
We define the partial order relation for dtypes used in our work:
\begin{equation}
float32 > int32/float16 > int16 > int8 > int4 > bool \nonumber
\end{equation}
The lower dtype can be converted into a higher dtype without accuracy loss, while a backward conversion with accuracy loss is forbidden.
Using lower dtype computation, such as int8 matmul, can speed up and reduce memory usage.
However, there are many limitations to dtype optimization.
For example, the inputs of the same operator should have the same dtype; thus, the dtype of operators depends on the largest dtype of inputs.
Besides, many hardware devices have extended instructions based on specific dtypes.
For example, an Intel processor speeds up int8 computation using AVX instruction, while bool cannot benefit from that.
Considering the complexity of dtype optimization, we add dtype as a property for ECG.

Sparsity is defined as the ratio of non-zero data to all data.
If data sparsity is relatively small, we take it as sparse data and store it in a compressed sparse row (CSR) format.
Using sparse operators to handle those sparse data can perform better than dense operators.
Taking advantage of sparsity influences optimization greatly, so we add sparsity as another property for ECG.

We classify the inputs of an operator into two categories: intermediate results and weights.
Intermediate results are other operators' outputs and can only be handled during runtime.
Input data is the first intermediate result in ECG, while output data is the last.
Intermediate results are represented as $\{sparsity,\ dtype,\ tensor\}$.
If we want to change the dtype of intermediate results, we should add dtype converting operator in the ECG.

Weights are model parameters that can be loaded from trained models.
Weights can be handled both during compilation and runtime, while a proper transformation during compilation can reduce runtime costs.
Weights are represented as $\{sparsity,\ smallest\_dty-pe
, actual\_dtype, \ tensor\}$.
Smallest\_dtype is the smallest dtype for weights without accuracy loss, actual\_dtype is the dtype actually used. 
Smallest\_dtype depends on the property of weights, while actual\_dtype is fixed based on smallest\_dtype and operators.
As shown in Fig.~\ref{conversion}, $W_1$ represents the relationship between input features and internal nodes for decision trees, which is a 0-1 matrix.
The smallest\_dtype of $W_1$ is bool.
However, W1 is multiplied by input data with a dtype of float32.
If we choose bool as the actual\_dtype, $W_1$ will be converted to float32 during runtime.
To reduce the execution time in runtime, we should convert $W_1$ to float32 during compilation, so we set actual\_dtype as float32 rather than bool.

\begin{table}
	\caption{Operators used in ECGs}
	\label{operator}
	\center
	{
		\begin{tabular}
			{c|c}
			\hline
			Operator Type & Examples\\
			\hline
			Comparison& less, equal, greater, less\_equal\\
			\hline
			Indices& argmax, argmin, argsort, argwhere\\
			\hline
			Monotonic& sigmoid, softmax, relu, tanh, exp\\	
			\hline	
			Reduction& sum, max, min, avg, all, any\\
			\hline
			Arithmetic& gemm, conv, pool\\	
			\hline		
	\end{tabular}}
\end{table}

Operators are represented in the form of  $\{weights,\ intermediate\_ \\
results,\ use\_sparse,\ type,\ dtype,\ DL\_operator\}$.
Weights and intermediate\_results are inputs of operators.
Use\_sparse is a flag of whether using the sparse operator or not, which is closely related to sparse operator replacing optimization described in Section~\ref{section_sparse_operator_replacing}.
Operator type is the type of operator.
As shown in Table~\ref{operator}, we divide operators used in ECG into five categories.
Comparison operators refer to those operators that compare two tensors and return bool tensors.
Indices operators refer to those operators that return tensors' indices based on specific conditions.
Those two kinds of operators are dtype-lowering operators, the output dtype of which is smaller than the input.
Models without those operators, such as most DL models, use the same dtype through the whole graphs, where dtype optimizations cannot be used without approximate optimization.
CML models make much use of those operators, which have wide usage of dtype rewriting optimization described in Section~\ref{section_dtype_rewriting}.
Monotonic operators refer to those operators who meet the following conditions:
\begin{equation}
\forall x_1 \leq x_2 \implies f(x_1) \leq f(x_2) \nonumber
\end{equation}
A series of monotonic operators followed by an indices operator is mathematically equivalent to the indices operators alone.
Those properties provide more optimizations, as described in Section~\ref{section_redundant_elimination}.
Reduction operators calculate aggregates over input.
Arithmetic operators refer to other arithmetic calculations.
Operator dtype is the operators' data type, such as int8 matmul or float32 matmul.
Operator dtype depends on the dtype of weights and intermediate\_results.
DL\_operator is the native definition of operators in DL computational graphs, which we use to translate ECG to DL computational graphs.
\subsection{Supported Algorithms and Extension for Other Algorithms}\label{section_supported}
\begin{table}
	\caption{Supported Algorithms}
	\label{supported_algorithms}
	\center
	{
		\begin{tabular}
			{|c|}
			\hline
			\textbf{Preprocessing Algorithms}\\
			\hline
			Binarizer, LabelBinarizer, Normalizer, MaxAbsScaler,\\ MinMaxScaler,
			StandardScaler, RobustScaler,\\
			PolynomialFeatures, LabelEncoder\\
			\hline
			\textbf{Feature Selectors}\\
			\hline
			SelectKBest, VarianceThreshold\\
			\hline
			\textbf{Linear Models}\\
			\hline
			LogisticRegression, LogisticRegressionCV, Perception, \\
			RidgeClassifier, RidgeClassifierCV, SGDClassifier, \\
			LinearRegression, Ridge, RidgeCV, SGDRegressor \\
			\hline
			\textbf{Tree-based Models}\\
			\hline
			DecisionTreeClassifier, DecisionTreeRegressor,\\ ExtraTreeClassifier, 
			ExtraTreeRegressor, \\
			RandomForestClassifier, RandomForestRegressor,\\
			ExtraTreesClassifier, ExtraTreesRegressor, \\
			GradientBoostingClassifier,
            GradientBoostingRegressor\\
			\hline
			\textbf{Support Vector Machines}\\
			\hline
			LinearSVC, LinearSVR, NuSVR, SVR \\
			\hline
	\end{tabular}}
\end{table}
CMLCompiler supports 35 CML algorithms nowadays, as shown in Table \ref{supported_algorithms}, covering most of the popular CML algorithms~\cite{ray2019quick}.
Our work can also be extended to other algorithms, such as clustering and matrix decomposition.
Most CML algorithms use operators categorized in Section ~\ref{section_conversion_category}, each of which can be converted to corresponding Operator Representations---our low-level abstractions, guaranteeing our extensibility.
We take Kmeans as an example.
Kmeans use basic arithmetic operators to calculate the distance between nodes, which can be converted to element-wise arithmetic operators and use aggregation operators to make clustering, which can be converted to reduction operators.
When all operators of a CML algorithm are converted to Operator Representations, it can utilize our work to compile and make optimizations.
\section{Design and Implementation}\label{section_design}
\begin{figure}
	\centering
	\includegraphics[width=0.48\textwidth]{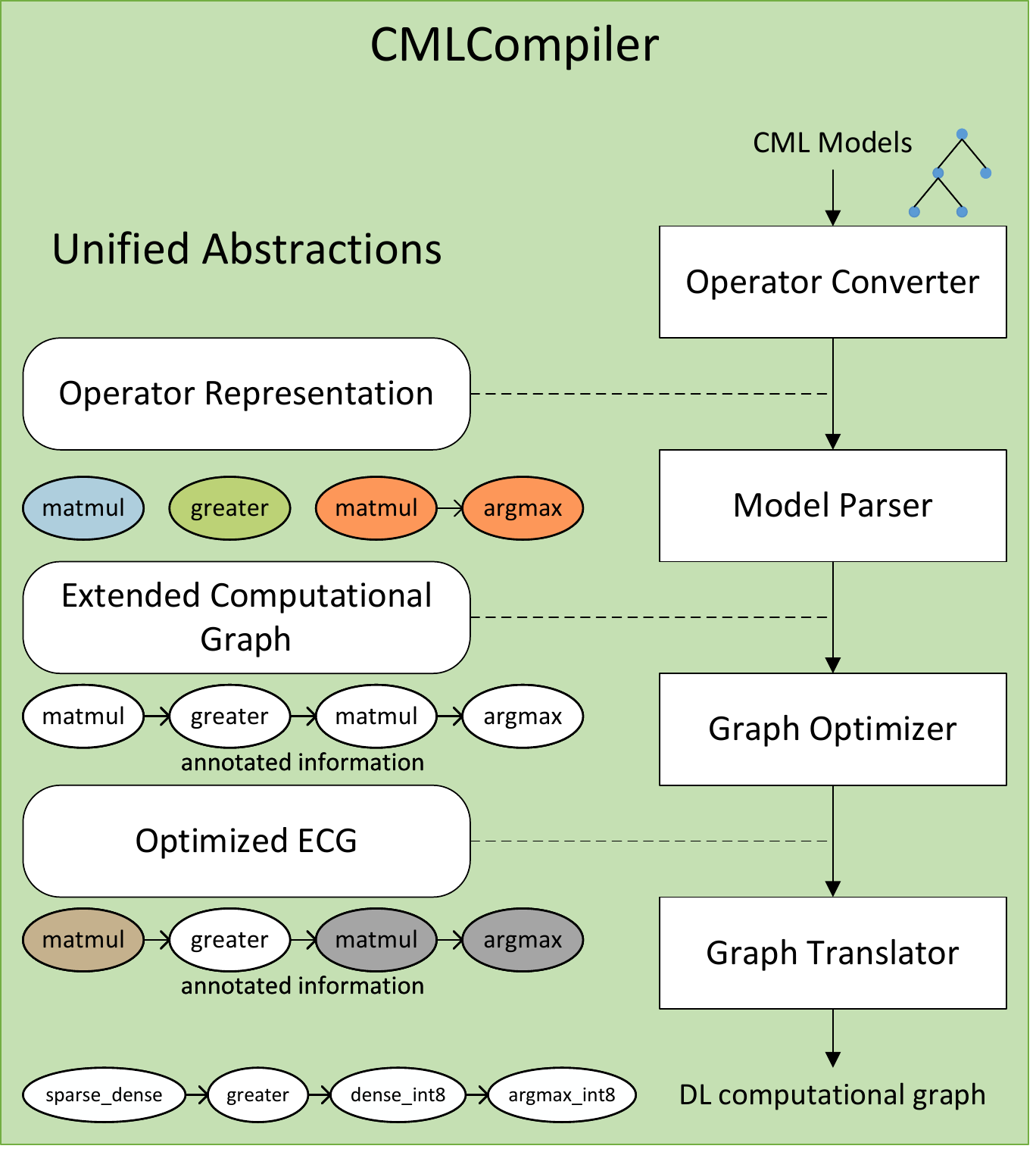}\\
	\caption{The CMLCompiler architecture.}
	\label{framework}
\end{figure}

This section illustrates the design and implementation of CMLCompiler, as shown in Fig.~\ref{framework}.
We build our framework based on the two unified abstractions, including four parts.
Operator Converter converts CML operators into operator representations, as shown in Section~\ref{section_operator_converter}.
Model Parser organizes those operator representations in an optimization-friendly way and uses ECGs to represent CML models, as shown in  Section~\ref{section_model_converter}.
Graph Optimizer makes graph level optimizations, as described in Section~\ref{section_graph}.
An optimized ECG is converted into a DL computational graph by Graph Translator in Section~\ref{section_translator}.
DL frameworks or compilers take DL computational graphs as input and make more optimizations, compiling them into executable modules to deploy.
Section~\ref{section_mixture} shows the mixture usage of CML and DL.
Section~\ref{section_implementation} shows the implementation details.

\subsection{Operator Converter}\label{section_operator_converter}
Operator Converter traverses the operators in CML models and converts them into operator representations, respectively.
If CML operators handle matrices or arrays, they can be converted to DL operators directly. If CML operators handle scalars or a single element from data tables, they are converted to DL operators based on their categories. Several assignment operators without data dependencies are converted to one tensor assignment. If the data processing order differs from its layout in the tensor, we use the $reorganization$ operator to change the tensor layout. Basic arithmetic operators are converted to element-wise arithmetic operators. Aggregation operators are converted to reduction operators. Comparison operators are converted to element-wise comparison. Sequential conditional operators are represented as the combination of $matmul$ and $argmax$. These converted DL operators are wrapped into operator representations with data dependencies.

\subsection{Model Parser}\label{section_model_converter}
Model Parser converts operator representations into an ECG.
Operators in an operator representation are initialized as nodes in an ECG, the data structure of which is defined in Section~\ref{section_ECG}.
Operator.weights and operator.intermediate\_results are set according to data dependencies, and edges are built between nodes.
Operator.use\_sparse and operator.dtype are set as False and Unknown, respectively.
Operator.type is set according to operator type, which is defined in Table~\ref{operator}.
Weights and intermediate\_result are initialized after that.
Weight.sparsity is set as the ratio of non-zero data and all data, known during compilation.
Weight.smallest\_dtype is set as the smallest dtype without accuracy loss, and weight.actual\_dtype is initialized the same.
Intermediate\_result.sparsity and intermediate\_result.dtype are set according to operator.
When all operators are visited, the ECG is established.

\subsection{Graph Optimizer}\label{section_graph}

Graph Optimizer performs graph-level optimizations, using a functionally equivalent transformation for ECGs. 
These optimizations are based on the features of CML models and do not influence accuracy.
There are three specific graph rewriting optimizations: dtype rewriting, sparse operator replacing, and redundant elimination.

\begin{figure*}
\begin{subfigure}{0.31\textwidth}
  \centering
  \includegraphics[width=1.0\linewidth]{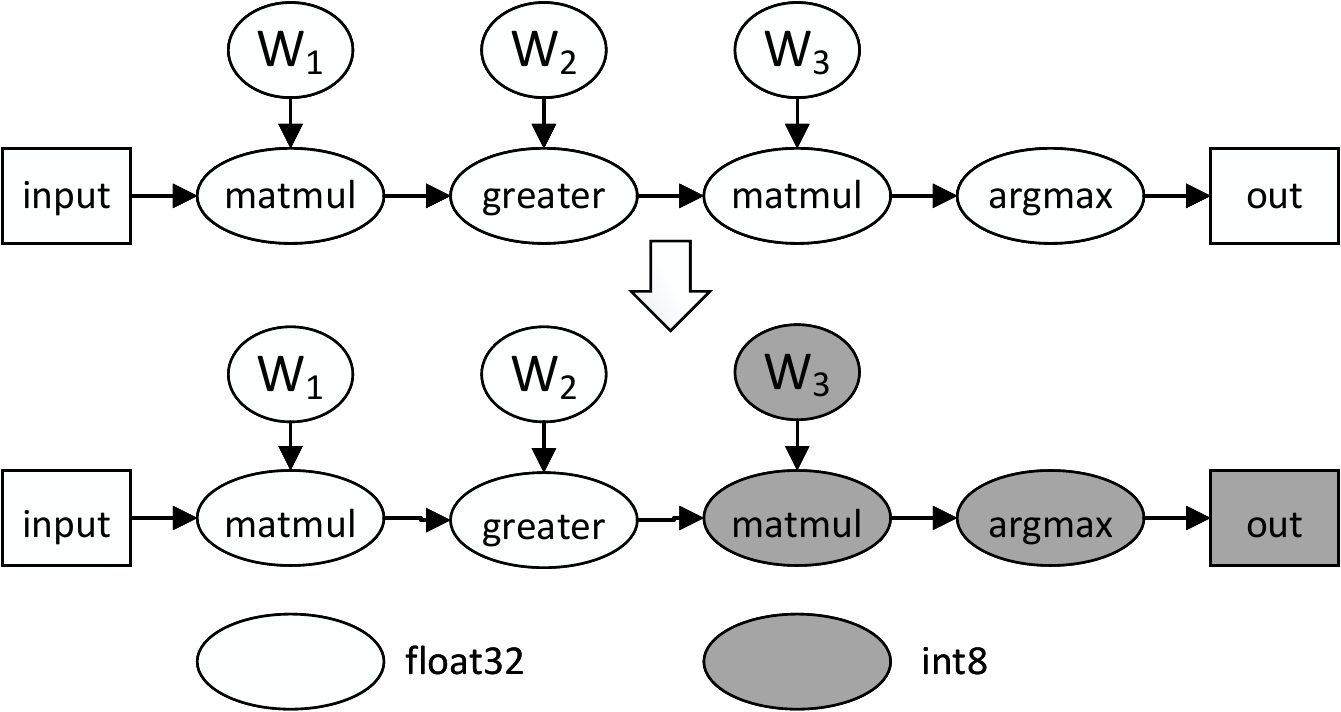}  
  \caption{Dtype Rewriting}
  \label{fig_dtype_rewrting}
\end{subfigure}
\hspace{1em}
\begin{subfigure}{0.31\textwidth}
  \centering
  \includegraphics[width=1.0\linewidth]{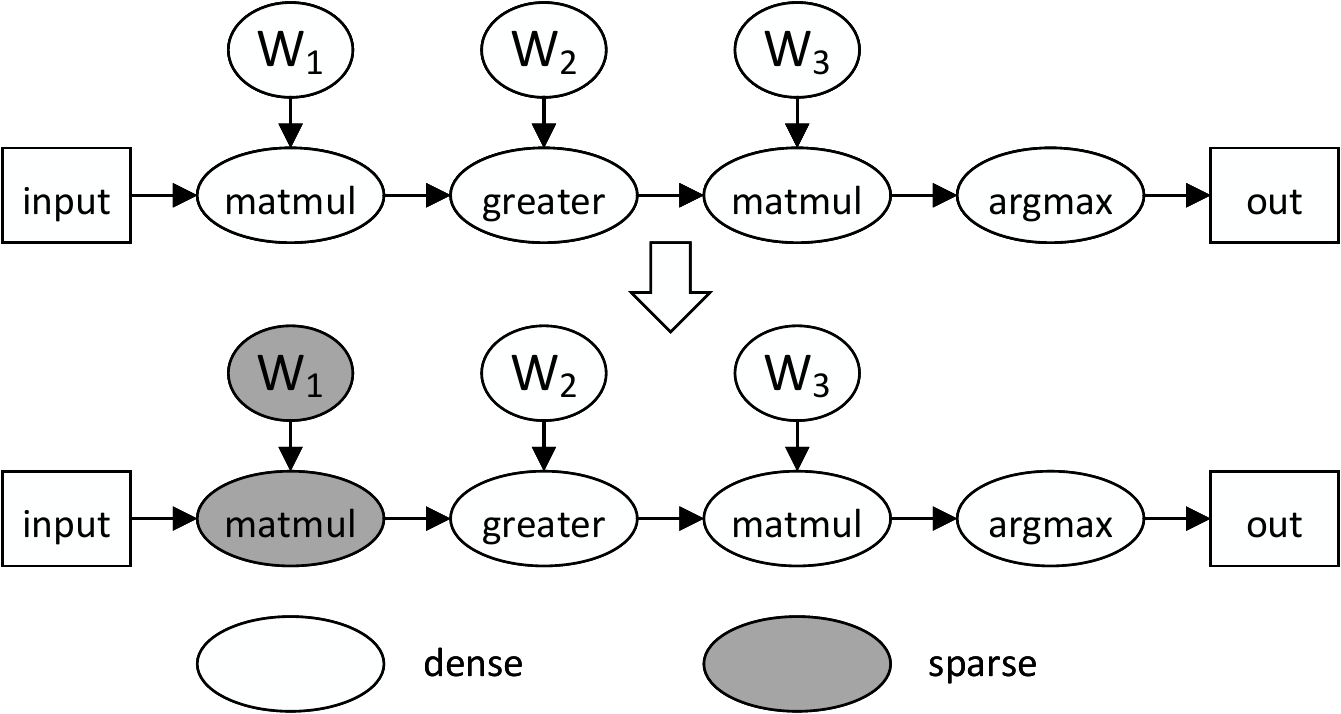}  
  \caption{Sparse Operator Replacing}
  \label{fig_sparse_operator_replacing}
\end{subfigure}
\hspace{1em}
\begin{subfigure}{0.31\textwidth}
  \centering
  \includegraphics[width=1.0\linewidth]{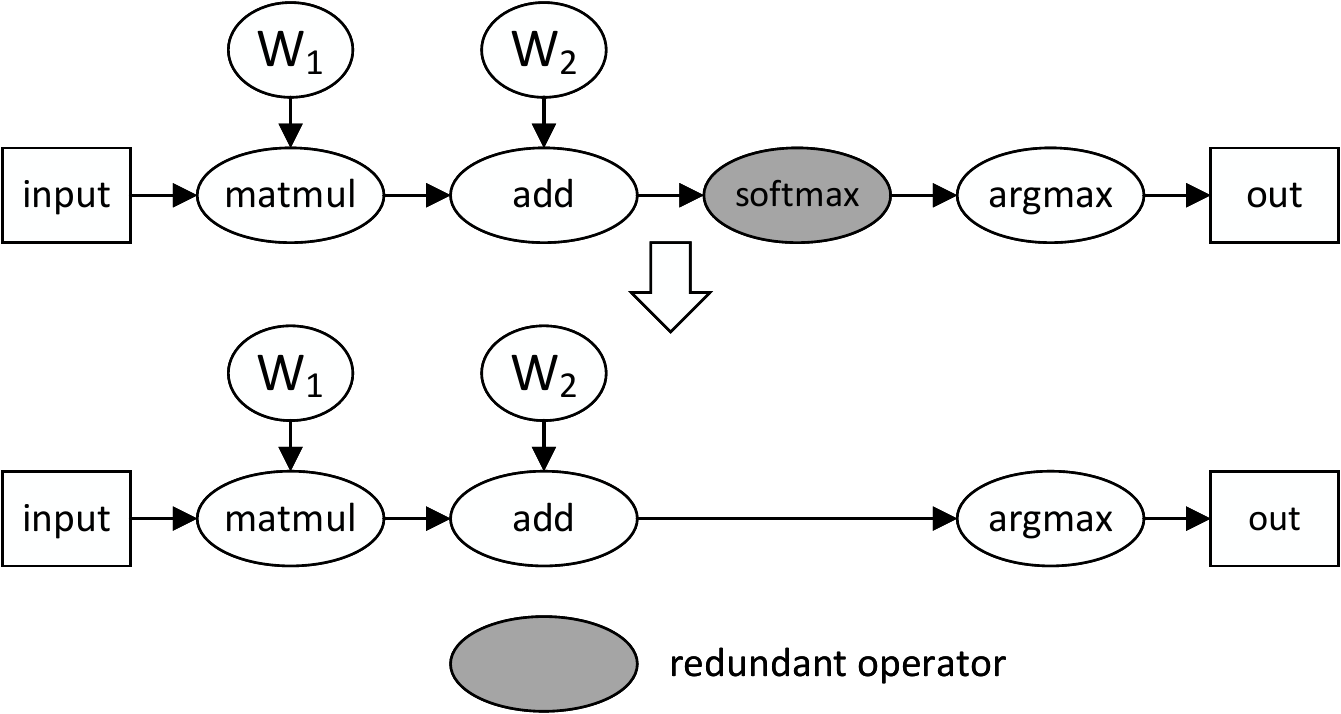}  
  \caption{Redundant Elimination}
  \label{fig_redundant_elimination}
\end{subfigure}
\caption{Graph rewriting optimizations. Dtype rewriting converts float32 operators and weights into low-precision. Sparse operator replacing converts dense operators and weights into sparse. Redundant elimination reduces redundant operators.}
\label{fig_graph_optimization}
\end{figure*}

\subsubsection{Dtype rewriting.}\label{section_dtype_rewriting}
Dtype rewriting uses low-precision computation with faster speed and less memory to replace high-precision computation.
As analyzed in Section~\ref{section_feature}, many weights used in CML can be represented as low-precision dtype such as bool or int8.
Besides, comparison operators and indices operators widely used in CML are dtype-lowering operators. 
The intermediate results after those operators are also low-precision.
When intermediate data and weights can both be expressed as low-precision dtype, the corresponding operators can be converted into low-precision computation as well.

As shown in Fig.~\ref{fig_dtype_rewrting}, the top is the ECG of decision trees before optimization; many details are hidden.
Weight $W_3$ represents the relationship between leaf nodes and internal nodes for decision trees, which is a matrix only containing 0 and 1.
The smallest\_dtype of $W_3$ is bool.
The output of $greater$ operator has a dtype of bool as well.
So the following matrix multiplication (matmul) operator can use a dtype of bool rather than float32.
Intel processors speed up int8 computation using AVX instruction, while bool cannot benefit from that feature.
So we convert the dtype of matmul to int8 according to hardware specification.
In Fig.~\ref{fig_dtype_rewrting}, below is the ECG after graph rewriting.
Those white weights and operators use float32, while gray weights and operators use int8.

Now we introduce the dtype rewriting principle in detail.
Algorithm \ref{dtype_algo} shows the procedure of dtype rewriting:

(1) Visit all operators in ECG.
For each operator, dtype is set as the largest dtype of all inputs.
After that, operator dtype is converted to the dtype which can utilize hardware’s SIMD
instructions best.
We keep a list of hardware specifications to modulate operator dtype.
In order to guarantee accuracy, dtype cannot get smaller.
Then we modulate operator implementation based on operator dtype.

(2) When operator dtype is fixed, we set the input dtype.
The dtype of weights is set the same as the operator, reducing dtype conversion in runtime.
The dtype of intermediate results cannot be converted during compilation.
So we add dtype converting operator, .i.e, cast, before the operator.

We explain the differences between dtype rewriting for CML models and model quantization for DL models.
Quantization is an approximate algorithm for DL models that causes a decrease in accuracy and brings extra computation, such as calibration.
Dtype rewriting for CML models is based on the properties of CML, converting dtype of operators and weights with no accuracy decrease and extra computation.

\begin{algorithm}
	\caption{Dtype Rewriting}\label{dtype_algo}
	\begin{algorithmic}[0]
		\REQUIRE ECG $G$, hardware configuration $H$
		\ENSURE Optimized ECG $G'$
		\FOR {operator in $G$}	
			\STATE{operator.dtype $\gets$ largest dtype in operator.weights and operator.intermediate\_results}		
			\STATE{Modulate operator.dtype based on $H$}
			\STATE{Modulate operator.DL\_operator based on operator.dtype}
			\FOR {weight in operator.weights}		
				\STATE{weight.actual\_dtype $\gets$ operator.dtype}		
			\ENDFOR
			\FOR {data in operator.intermediate\_results}
				\IF {data.dtype $<$ operator.dtype}
					\STATE{Add cast(data, operator.dtype) before operator}
				\ENDIF
			\ENDFOR
		\ENDFOR
	\end{algorithmic} 
\end{algorithm}

\subsubsection{Sparse operator 
replacing.}\label{section_sparse_operator_replacing}
Replacing dense operators with sparse operations can speed up as well.
The sparsity of input data can be known until runtime, while the sparsity of weights can be known during compilation.
So we convert the data format of weights rather than input data.
Different hardware devices have different support for sparse operators.
For example, CPUs can benefit from sparse computation while GPUs have little effect.
So we set a threshold based on hardware specification.
If weight.sparsity is smaller than the threshold, we store it in a compressed sparse row (CSR) format.
Then we convert the corresponding operator into a sparse implementation.
An example is shown in Fig.~\ref{fig_sparse_operator_replacing}, we convert $W_1$ and the corresponding matmul to sparse.

\subsubsection{Redundant elimination.}\label{section_redundant_elimination}

Redundant elimination eliminates those operators who do not influence final results due to their mathematical properties.
For example, a series of monotonic operators followed by an indices operator is mathematically equivalent to the indices operators alone.
For each operator in ECGs, we check its operator type.
If another monotonic operator follows a monotonic operator, we fuse them.
We eliminate the monotonic operator if an indices operator follows it.
An example is shown in Fig.~\ref{fig_redundant_elimination}, the softmax before argmax is eliminated.

\subsection{Graph Translator}\label{section_translator}
Graph Translator converts the optimized ECG into DL computational graph based on ECG topology and chooses the proper operator implementation.
DL frameworks or compilers provide different implementations for the same operator.
Graph Translator utilizes four properties: $use\_sparse$, $type$, $dtype$, and $DL\_operator$ in ECG to choose the most proper implementation.
Operator implementation can also be modulated based on hardware information.
We can utilize hardware features to optimize operators.
If the hardware supports extended instructions like AVX, we use them to speed up operators.
If the backend is TVM, we just pass the precise $mcpu$ and $mattr$ information and utilize TVM to make operator-level optimizations.
DL frameworks or compilers take DL computational graphs as input and make more optimizations, finally compiling them into executable modules.

\subsection{Hybrid Deployment of CML and DL with a Unified Framework}\label{section_mixture}
\begin{figure}
	\centering
	\includegraphics[width=0.48\textwidth]{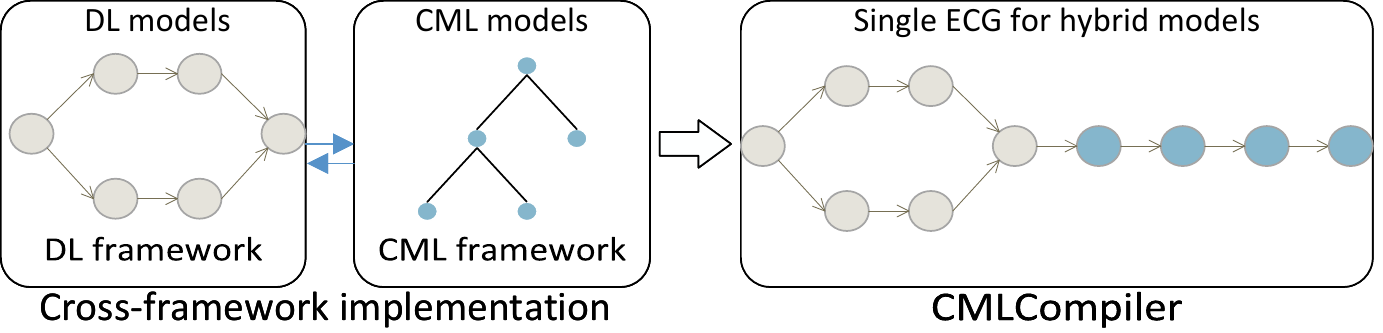}\\
	\caption{CMLCompiler uses a single ECG to represent CML and DL mixed pipeline.}
	\label{mixture}
\end{figure}

We convert those CML and DL hybrid applications under a unified framework to reduce the cost of switching frameworks and provide an opportunity for end-to-end optimizations, as shown in Fig.~\ref{mixture}.
We load models from PyTorch and sklearn and convert them into ECG subgraphs.
We build edges according to data dependency and merge those subgraphs in a single ECG.
Then we can use optimizations both in our work and DL compilers.
Finally, we compile and deploy it on various hardware devices.

\subsection{Implementation}\label{section_implementation}
Due to the benefits in portability and performance, we implement CMLCompiler on the basis of TVM.
The intermediate representations and transforms are all written in python.
We read trained models from CML frameworks such as sklearn and convert them into operator representations, implementing them in the format of TVM relay functions and storing their weights in TVM arrays.
We wrap those relay functions in the format of ECGs.
After optimizations in Section~\ref{section_graph}, we convert ECGs into TVM's IRModules.
Then we utilize TVM to make more optimizations and compile to executable modules based on specific hardware targets.
We use cross-compilation to support a broad spectrum of hardware devices.
We deploy them on lightweight runtime based on TVM runtime and make inferences on various hardware devices.

\section{Evaluation}\label{section_evaluation}
This section summarizes the evaluation.
Section~\ref{sec_setup} shows the experimental setup.
Section~\ref{sec_optimization} evaluates the performance of graph rewriting optimizations based on ECGs.
Section~\ref{sec_overall} compares our work with the  state-of-the-art frameworks.
Section~\ref{section_mix} evaluates the hybrid deployment of CML and DL.
\subsection{Experimental Setup}\label{sec_setup}

We deploy a server node equipped with two Xeon E5-2620 V3 (Haswell) CPUs, an Nvidia Titan RTX GPU, and 64 GB memory to conduct the experiments on CPU and GPU. 
Each CPU contains six physical cores.
The GPU contains 4608 Cuda cores and 24 GB memory.
The operating system is Ubuntu 16.04, and the other software includes TVM 0.8, PyTorch 1.8.1, hummingbird 0.3.1, scikit-learn 1.0.1, and CUDA 10.2.
For the IoT experiments, we use Raspberrypi4b with Raspbian 10 operating system and deploy the above software with the same version.
We use YearPrediction~\cite{Dua:2019} as the dataset, with 515345 samples and 90 features. 
We use 80\% data to train models and 20\% data to make inferences.
We run all the experiments five times and use the average as the final results.
We test hummingbird~\cite{nakandala2020tensor} using both two backends (PyTorch and TVM) and select their best results.

\subsection{Optimizations}\label{sec_optimization}

\begin{figure*}
\begin{subfigure}{0.49\textwidth}
  \centering
  \includegraphics[width=1.0\linewidth]{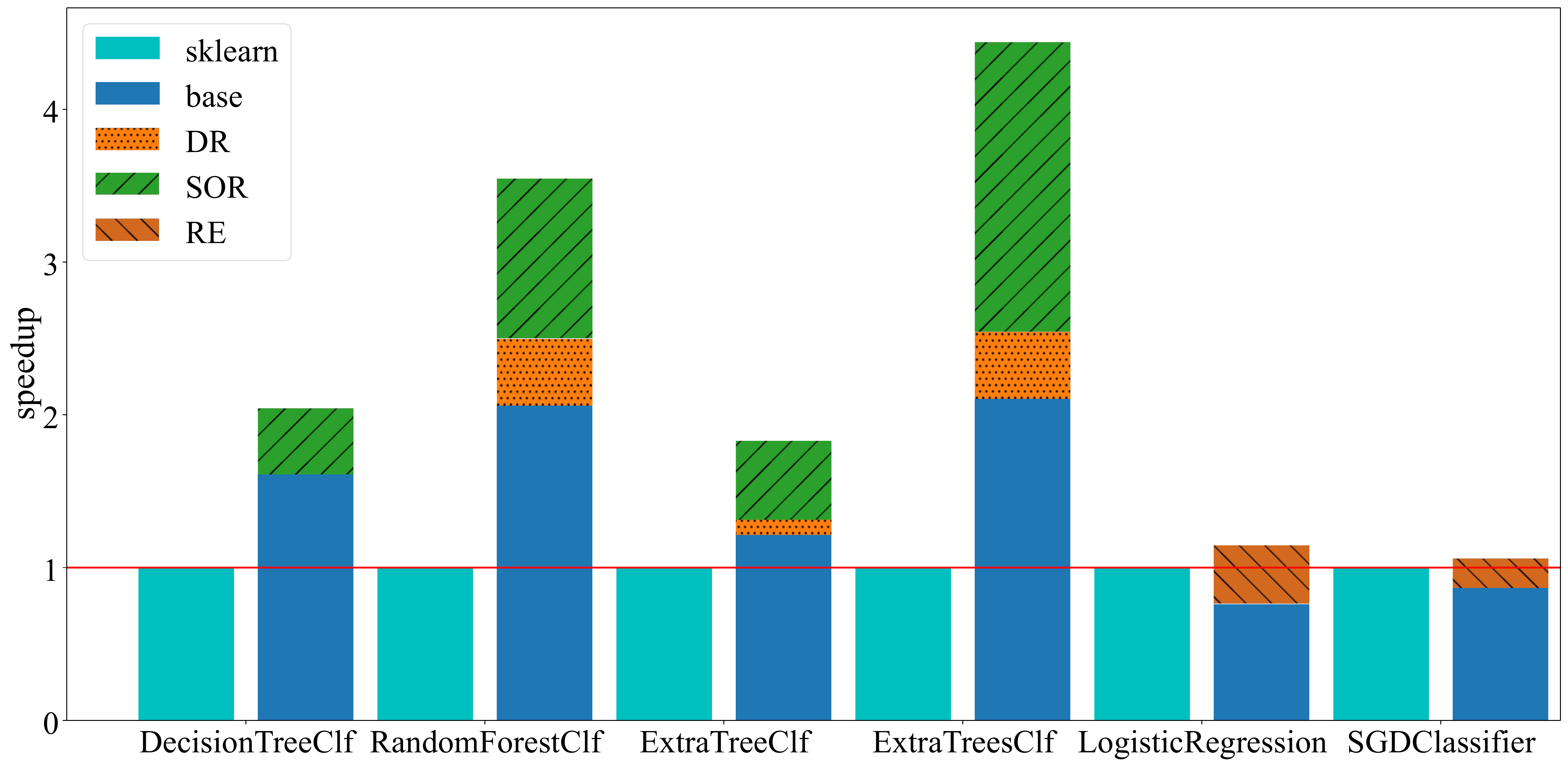}  
  \caption{CPU}
  \label{cpu_optimization}
\end{subfigure}
\begin{subfigure}{0.49\textwidth}
  \centering
  \includegraphics[width=1.0\linewidth]{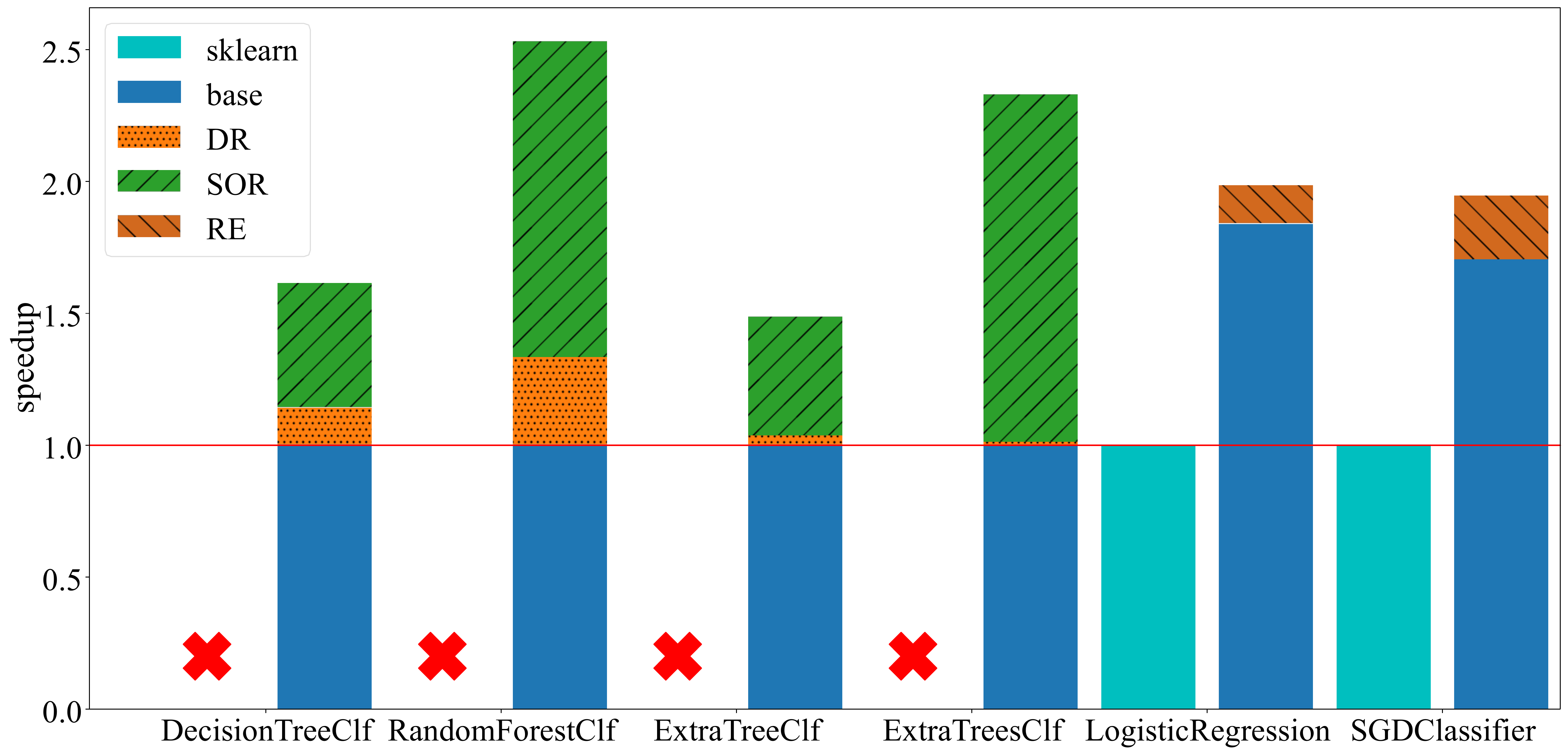}  
  \caption{Raspberrypi4b}
  \label{iot_optimization}
\end{subfigure}
\caption{Graph Rewriting Optimizations. 
        Take sklearn as the baseline.
	"base" means our work without optimizations.
	"DR" means using dtype rewriting.
	"SOR" means using sparse operator replacing.
	"RE" means using redundant elimination.
        "$\times$" means unsupported.
        For those models not supported by sklearn on IoT devices, take our result without optimzations as the baseline.
        }
\label{optimization}
\end{figure*}

This section evaluates graph rewriting optimizations based on ECGs, as described in Section~\ref{section_graph}.
These optimizations: dtype rewriting, sparse operator replacing, and redundant elimination, can work together and produce cumulative optimization effects.
They can also coexist with the optimizations in TVM.
We choose four typical tree models: DecisionTreeClassifier, RandomForestClassifier, ExtraTreeClassifier, and ExtraTreesClassifier, as well as two typical linear models: LogisticRegression and SGDClassifier. 
We evaluate the dtype rewriting and sparse operator replacing for tree models, and redundant elimination for linear models according to their unique patterns. 

Fig.~\ref{cpu_optimization} shows the result on CPU.
For tree models, using our work without optimizations has a 1.31x-2.54x speedup compared with sklearn; this is due to our abstractions which utilize optimizations of TVM, including better utilization of SIMD instructions and multi cores.
Using dtype rewriting and sparse operator replacing bring 1x-1.21x and 1.26x-1.75x speedup, respectively, achieving 1.27x-2.11x speedup together, 1.84x-4.44x faster than sklearn.
For linear models, our work without optimizations runs slower than sklearn.
However, using redundant elimination brings 1.22x-1.51x speedup; the result after our optimizations is 1.06x-1.14x faster than sklearn.

Fig.~\ref{iot_optimization} shows the result of IoT devices.
Note that sklearn lacks enough support for IoT devices.
For example, 64-bit tree models trained on servers cannot be executed on Raspberrypi4b with a 32-bit operating system.
Retraining those models in 32-bit format on Raspberrypi4b from scratch takes more time, so we regard those models as unsupported, marked as cross.
So we take our work without optimizations as the baseline.
Using dtype rewriting and sparse operator replacing bring 1.01x-1.33x and 1.23x-2.3x speedup, respectively, achieving 1.49x-2.53x speedup together.
For linear models, our work without optimizations achieves a 1.71x-1.84x speedup.
Using redundant elimination brings 1.08x-1.14x more speedup, 1.95x-1.98x faster than sklearn.
The computation part of GPU is less than 20\%, so those optimizations play a limited role on GPU.
In conclusion, CML models can benefit from both TVM's optimizations and our optimizations and achieve obvious speedup.

\subsection{Overall Results}\label{sec_overall}

\begin{table*}
	\caption{Execution time for batch experiments over all data on CPU (12 cores), GPU, and IoT devices (take Raspberrypi4b as an example) in milliseconds. SK, HB, and Intel is short for scikit-learn, hummingbird, and Intel extension for sklearn, respectively. 
	"-" means unsupported.}
	\label{overall_result}
	\center
	{
		\begin{tabular}
			{p{1.6in}<{\centering}p{0.4in}<{\centering}p{0.4in}<{\centering}p{0.4in}<{\centering}p{0.4in}<{\centering}p{0.4in}<{\centering}p{0.4in}<{\centering}p{0.4in}<{\centering}p{0.4in}<{\centering}}
			\toprule 
			\multirow{2}[2]{*}{Algorithm}&\multicolumn{4}{c}{CPU}&\multicolumn{2}{c}{GPU}&\multicolumn{2}{c}{IOT}\\
			\cmidrule(lr){2-5} \cmidrule(lr){6-7} \cmidrule(lr){8-9}
			&SK &HB &Intel &Our & HB &Our &SK &Our\\
			\midrule
			Binarizer & 97 & 31 & 77 & \textbf{9} & 19 & \textbf{6} & 634 & \textbf{126}\\
                Normalizer & 25 & 33 & \textbf{15} & \textbf{15} & 7 & \textbf{5} & 241 & \textbf{168}\\
                MinMaxScaler & 19 & 31 & 13 & \textbf{8} & 21 & \textbf{6} & 199 & \textbf{148}\\
                RobustScaler & 28 & 32 & 25 & \textbf{12} & 19 & \textbf{5} & 343 & \textbf{156}\\
                LinearRegression & 12 & 18 & \textbf{4} & 6 & \textbf{6} & 7 & \textbf{61} & 116\\
                LogisticRegression & 98 & 104 & 137 & \textbf{86} & \textbf{7} & \textbf{7} & 1889 & \textbf{952}\\
                SGDClassifier & 94 & 98 & 139 & \textbf{88} & 9 & \textbf{7} & 1886 & \textbf{969}\\
                DecisionTreeClassifier & 33 & 48 & 23 & \textbf{16} & 7 & \textbf{5} & - & \textbf{99}\\
                DecisionTreeRegressor & 7 & 19 & \textbf{3} & 15 & 7 & \textbf{6} & - & \textbf{211}\\
                RandomForestClassifier & 2130 & 885 & 2003 & \textbf{601} & \textbf{20} & - & - & \textbf{5820}\\
                ExtraTreeClassifier & 29 & - & 26 & \textbf{16} & - & \textbf{6} & - & \textbf{206}\\
                ExtraTreesClassifier & 10022 & 2522 & 9421 & \textbf{2256} & \textbf{99} & - & - & \textbf{47959}\\
                LinearSVC & 92 & 122 & 152 & \textbf{77} & 9 & \textbf{6} & 1896 & \textbf{930}\\
                LinearSVR & 39 & 26 & 34 & \textbf{5} & 6 & \textbf{5} & 323 & \textbf{112}\\
			\bottomrule
	\end{tabular}}
\end{table*}

\begin{table*}
	\caption{Latency for query experiments over one single record on CPU (12 cores), GPU, and IoT devices (take Raspberrypi4b as an example) in milliseconds. The symbols are the same as Table~\ref{overall_result}.}
	\label{overall_latency}
	\center
	{
		\begin{tabular}
			{p{1.6in}<{\centering}p{0.4in}<{\centering}p{0.4in}<{\centering}p{0.4in}<{\centering}p{0.4in}<{\centering}p{0.4in}<{\centering}p{0.4in}<{\centering}p{0.4in}<{\centering}p{0.4in}<{\centering}}
			\toprule 
			\multirow{2}[2]{*}{Algorithm}&\multicolumn{4}{c}{CPU}&\multicolumn{2}{c}{GPU}&\multicolumn{2}{c}{IOT}\\
			\cmidrule(lr){2-5} \cmidrule(lr){6-7} \cmidrule(lr){8-9}
			&SK &HB &Intel &Our & HB &Our &SK &Our\\
			\midrule
            Binarizer & 0.2 & 0.26 & 0.34 & \textbf{0.09} & 0.93 & \textbf{0.64} & \textbf{0.44} & 0.59\\
            Normalizer & 0.32 & 0.26 & 0.28 & \textbf{0.11} & \textbf{0.25} & 0.68 & 0.59 & \textbf{0.41}\\
            MinMaxScaler & 0.15 & 0.31 & 0.14 & \textbf{0.09} & 0.91 & \textbf{0.63} & \textbf{0.33} & 0.37\\
            RobustScaler & 0.14 & 0.22 & 0.14 & \textbf{0.11} & 1.02 & \textbf{0.72} & \textbf{0.37} & \textbf{0.37}\\
            LinearRegression & 0.24 & 0.35 & 0.32 & \textbf{0.1} & 0.91 & \textbf{0.55} & \textbf{0.52} & 0.69\\
            LogisticRegression & 0.35 & 0.36 & 0.29 & \textbf{0.19} & 3.29 & \textbf{0.71} & \textbf{0.67} & 2.59\\
            SGDClassifier & 0.4 & 0.35 & 0.29 & \textbf{0.23} & 2.93 & \textbf{0.67} & 0.68 & \textbf{0.65}\\
            DecisionTreeClassifier & \textbf{0.24} & 1.62 & 0.27 & 0.36 & 3.01 & \textbf{0.8} & - & \textbf{0.9}\\
            DecisionTreeRegressor & \textbf{0.22} & \textbf{0.22} & 0.25 & 0.38 & 1.03 & \textbf{0.72} & - & \textbf{0.88}\\
            RandomForestClassifier & 103.96 & 1.6 & 103.2 & \textbf{0.61} & \textbf{2.56} & - & - & \textbf{1.05}\\
            ExtraTreeClassifier & \textbf{0.23} & - & 0.4 & 0.47 & - & - & - & \textbf{1.81}\\
            ExtraTreesClassifier & 205.27 & 12.74 & 204.25 & \textbf{1.73} & \textbf{2.41} & - & - & \textbf{3.11}\\
            LinearSVC & 0.4 & 0.37 & 0.45 & \textbf{0.19} & 2.71 & \textbf{0.61} & \textbf{0.65} & 1.07\\
            LinearSVR & 0.31 & 0.34 & 0.37 & \textbf{0.09} & 0.91 & \textbf{0.62} & \textbf{0.54} & 0.91\\
			\bottomrule
	\end{tabular}}
\end{table*}

This section evaluates 14 typical CML algorithms covering preprocessing algorithms, linear models, tree-based models, and SVMs, on CPU, GPU, and IoT devices, compared with state-of-the-art frameworks including sklearn, intel extension for sklearn~\cite{intel_sklearn}, and hummingbird.
It contains two parts: batch experiments for all data and query experiments for a single record.

The differences between the results of CMLCompiler and sklearn are all less than $1\times10^{-5}$, which means that our work does not affect the accuracy.
The outputs on different hardware are all the same, so we focus on performance hereinafter.
Table~\ref{overall_result} shows the performance of batch experiments.
On CPU, our work reflects the best performance on 12 algorithms out of 14, achieving 1.02x-10.57x speedup compared with sklearn, 1.14x-4.38x speedup compared with hummingbird, and 1.44x-8.47x speedup compared with intel sklearn.
On GPU, our work achieves competitive performance compared with hummingbird. 
Our work performs better on 11 algorithms out of 14, with a 1.11x-3.31x speedup.
On an IoT device Raspberrypi4b,
our work performs better on 13 algorithms out of 14, with a 1.28x-5.09x speedup.

Table~\ref{overall_latency} shows the performance of query experiments for a single record.
On CPU, our work achieves the best performance on 11 algorithms out of 14, with a 1.36x-170.68x speedup compared with sklearn, a 1.56x-4.47x speedup compared with hummingbird, and a 1.31x-169.43x speedup compared with intel sklearn.
Our work has better performance on GPU on 10 algorithms out of 14 compared with hummingbird, with a 1.41x-4.64x speedup.
Our latency on Raspberrypi4b does not differ much compared with sklearn.
However, we perform better in model support.

In conclusion, we have advantages in both batch and query experiments for all three hardware devices.
Many models in sklearn only support a single core and cannot fully utilize the SIMD instructions.
We perform better than sklearn and intel sklearn due to better utilization of multi-cores and SIMD instructions through compilation.
Hummingbird uses both PyTorch and TVM as backends, where TVM performs better in most cases of our evaluations.
It implements models in PyTorch and converts them into TVM using $from\_pytorch$ API.
This conversion is not direct and efficient enough, causing a performance decrease.
Besides, hardware information is missed during conversion, which limits the optimizations of TVM for hummingbird.
We map ECGs into relay operators directly and select the most efficient implementation based on ECGs and hardware specification information.
Additionally, our abstractions bring more optimizations, as described in Section~\ref{section_graph}, bringing up to 2.53x speedup, working together to achieve better performance.

\subsection{Hybrid Deployment of CML and DL}\label{section_mix}
\begin{figure*}
\begin{subfigure}{0.32\textwidth}
  \centering
  \includegraphics[width=1.0\linewidth]{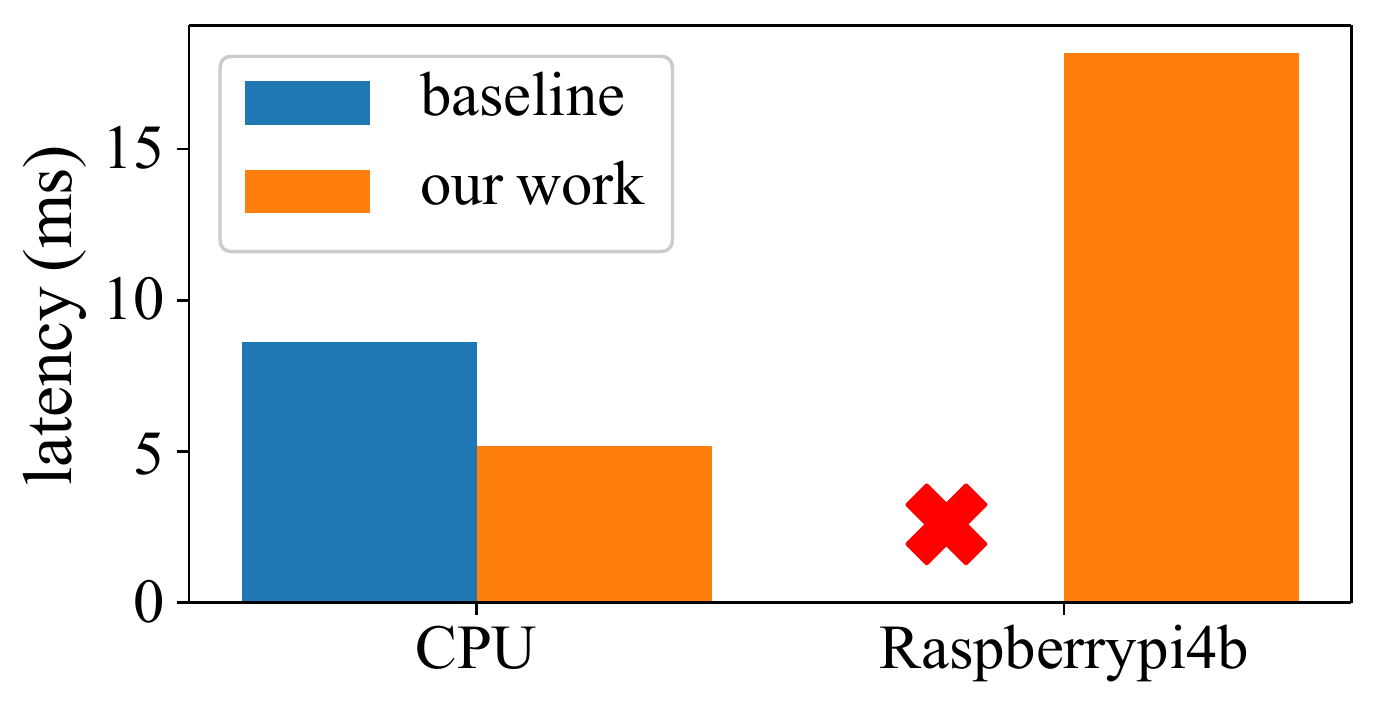}  
  \caption{\ \ Bert + LogisticRegression for sentence \\sentiment classification}
  \label{fig_mix_sentence}
\end{subfigure}
\begin{subfigure}{0.32\textwidth}
  \centering
  \includegraphics[width=1.0\linewidth]{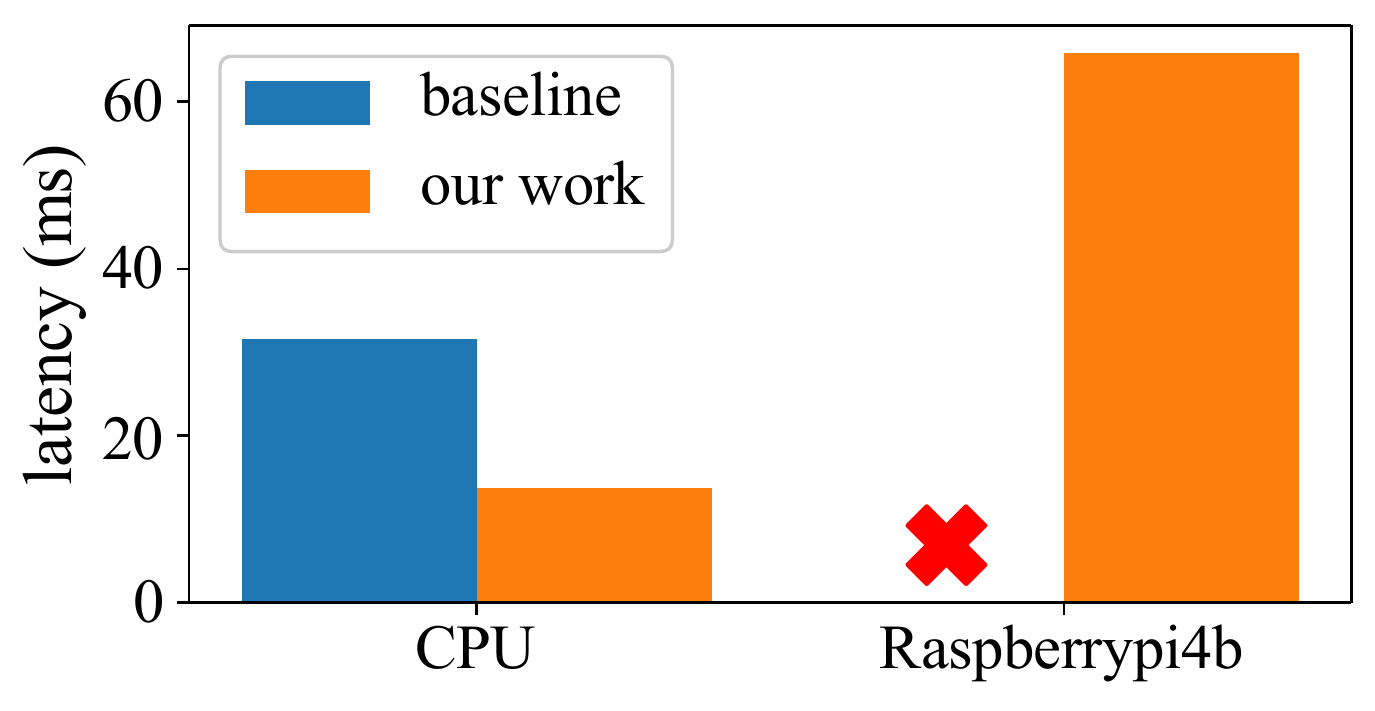}  
  \caption{\ \ SimpleDNN + RandomForest for \\radiographic image analysis}
  \label{fig_mix_radio}
\end{subfigure}
\begin{subfigure}{0.32\textwidth}
  \centering
  \includegraphics[width=1.0\linewidth]{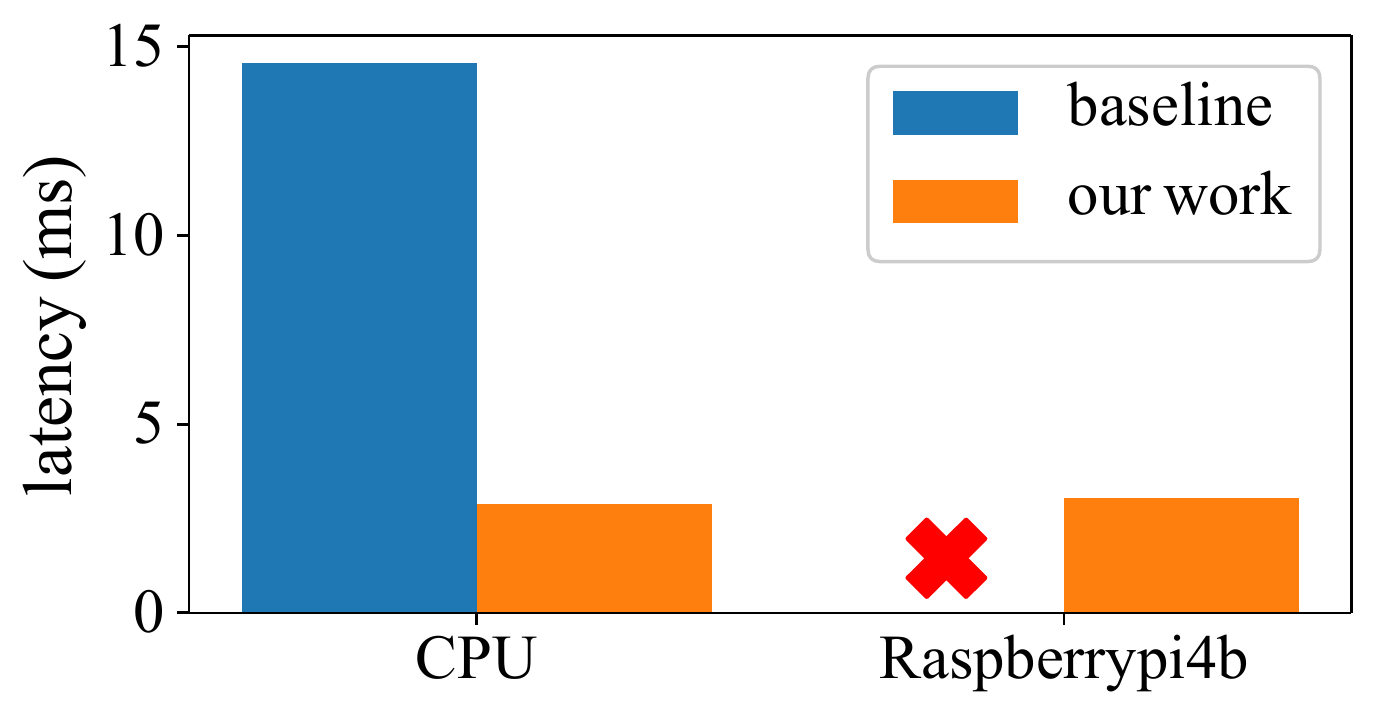}  
  \caption{\ \ GBDT + Wide\&Deep for click through \\prediction}
  \label{fig_mix_recommendation}
\end{subfigure}
\caption{The latency of a single query for CML and DL mixed pipelines. All three baselines cannot run on IoT devices.}
\label{fig_mix}
\end{figure*}
This section shows three hybrid deployment cases of CML and DL.
As the baselines, without a unified framework, a DL framework is used to implement DL algorithms, while a CML framework is used to implement CML algorithms. 
Our work converts CML and DL models into a single ECG, making optimizations and compiling them to various hardware devices.
We test the latency of a single query, which is essential in real-world applications.
\subsubsection{Sentence Sentiment Classification}
The first is a sentence sentiment classification case, which uses Bert to embed English sentences and logistic regression to make a classification~\cite{reimers2019sentence}.
We use BERT-tiny~\cite{bhargava2021generalization} as the pre-trained Bert model and SST2~\cite{socher2013parsing} as the dataset.
The baseline implements BERT-tiny in pytorch-transformers~\cite{wolf-etal-2020-transformers} and logistic regression in sklearn.
The result is shown in Fig~.\ref{fig_mix_sentence}.
Our work achieves a 1.67x speedup on server CPUs.
Pytorch-transformers cannot be installed on IoT devices, so the baseline cannot run on Raspberrypi4b.
The latency of our work on Raspberrypi4b is 18 milliseconds, which is acceptable in most use cases.
\subsubsection{Radiographic Image Analysis}
The second case uses Deep Hybrid Learning~\cite{sengupta2020nuclear} to analyze  radiographic images,
which uses simple DNN to make feature engineering and CML models such as random forests to make a classification.
We use CheXpert~\cite{irvin2019chexpert} as the dataset.
The baseline implements DNN in PyTorch and random forest in sklearn.
The result is shown in Fig~.\ref{fig_mix_radio}.
Our work achieves a 2.3x speedup on server CPUs.
The pre-trained random forest cannot run on IoT devices by sklearn, while our work can support those devices through cross-compilation.
\subsubsection{Click Through Rate Prediction}
The third case is click-through rate prediction used in recommendation systems of our anonymous industry partners, using GBDT~\cite{friedman2001greedy} to extract features and the Wide and Deep ~\cite{cheng2016wide} models to make a prediction.
We use avazu~\footnote{\url{https://www.kaggle.com/c/avazu-ctr-prediction}} as the dataset.
The baseline implements GBDT in sklearn and Wide and Deep in PyTorch.
The result is shown in Fig~.\ref{fig_mix_recommendation}.
We achieve 3.04x speedup on the server CPUs.
The GBDT model in the baseline cannot be executed on IoT devices, while our latency on IoT devices is only 5.06 ms.

\section{Related Work}\label{section_related_work}

CML frameworks and libraries can be divided into three categories.
(1) General-purpose solution uses one framework to support various models.
Scikit-learn~\cite{10.5555/1953048.2078195} is the most widely used CML framework on GitHub~\cite{DBLP:journals/corr/abs-1912-09536}.
Spark MLlib~\cite{10.5555/2946645.2946679} is an extension to Spark~\cite{zaharia2010spark}.
H2O~\cite{h2o_platform} uses MapReduce~\cite{dean2008mapreduce} to support both CML and DL.
There are many other works, such as Shogun~\cite{sonnenburg2010shogun} and RapidMiner~\cite{hofmann2016rapidminer}.
These frameworks only support CPU, suffering from severe performance and portability issues.
(2) Specific-purpose solution focuses on one type of model.
LibLinear~\cite{fan2008liblinear} supports logistic regression and linear SVM.
LibSVM~\cite{chang2011libsvm} focuses on SVMs.
These works are limited to CPUs.
Some other works attempt to support various hardware devices.
XGBoost~\cite{chen2016xgboost} implements a gradient-boosting decision tree algorithm on CPUs and GPUs.
Muhsen Owaida et al. ~\cite{owaida2017scalable} bring XGBoost to FPGAs.
Toby Sharp~\cite{sharp2008implementing} implements decision trees and forests on GPUs.
These frameworks only support a narrowed variety of models and solve the problem of portability to a certain extent. 
(3) Extension based on DL attempts to utilize DL frameworks to support CML models.
TF-DF~\cite{tf_df} is a decision forest library based on TensorFlow but is limited to CPUs.
It's implemented ad-hoc, losing the portability of DL frameworks.
Hummingbird~\cite{nakandala2020tensor} is a general-purpose solution based on PyTorch, adding support for GPUs.
They utilize those abstractions in DL frameworks directly without digging into the features of CML, missing many optimization chances.

\section{Conclusion}\label{section_conclusion}

This paper presented the design and implementation of CMLCompiler, an open-source unified compiler for classical Machine Learning (CML) inference.
CMLCompiler proposed two unified abstractions: operator representations and extended computational graphs (ECGs). Operator representations convert CML operators into tensor formats, while an ECG organizes these converted operators in an optimization-friendly way.
The CMLCompiler framework performs the conversion and graph optimization based on two unified abstractions, then outputs an optimized computational graph to deep learning compilers or frameworks. CMLCompiler also enables the hybrid deployment of CML and DL with a unified framework. Our implementations of CMLCompiler on top of TVM show the effectiveness and achieve up to 4.38x speedup on CPU, 3.31x speedup on GPU, and 5.09x speedup on IoT devices, compared to the state-of-the-art solutions --- scikit-learn, Intel sklearn, and hummingbird.
Our support for CML and DL mixed pipelines achieves up to 3.04x speedup compared with cross-framework implementations.

\balance
\bibliographystyle{plain}
\bibliography{references}
\end{document}